\documentclass[]{spie}  

\usepackage{amsmath,amsfonts,amssymb}
\usepackage{graphicx}
\usepackage[colorlinks=true, allcolors=blue]{hyperref}
\usepackage{makecell}

\title{Constraint-aware path planning from natural language instructions using large language models}

\author[a]{Dylan Shim}
\author[a]{Minghan Wei}
\affil[a]{Florida Atlantic University, 777 Glades Road, Boca Raton, United States}

\authorinfo{Further author information: (Send correspondence to D.S.)\\D.S.: E-mail: dshim2019@fau.edu\\  M.W.: E-mail: weim@fau.edu}

\begin{document} 
\maketitle

\begin{abstract}
Real-world path planning tasks typically involve multiple constraints beyond simple route optimization, such as the number of routes, maximum route length, depot locations, and task-specific requirements. Traditional approaches rely on dedicated formulations and algorithms for each problem variant, making them difficult to scale across diverse scenarios. In this work, we propose a flexible framework that leverages large language models (LLMs) to solve constrained path planning problems directly from natural language input. The core idea is to allow users to describe routing tasks conversationally, while enabling the LLM to interpret and solve the problem through solution verification and iterative refinement. The proposed method consists of two integrated components. For problem types that have been previously formulated and studied, the LLM first matches the input request to a known problem formulation in a library of pre-defined templates. For novel or unseen problem instances, the LLM autonomously infers a problem representation from the natural language description and constructs a suitable formulation in an in-context learning manner. In both cases, an iterative solution generation and verification process guides the LLM toward producing feasible and increasingly optimal solutions. Candidate solutions are compared and refined through multiple rounds of self-correction, inspired by genetic-algorithm-style refinement. We present the design, implementation, and evaluation of this LLM-based framework, demonstrating its capability to handle a variety of constrained path planning problems. This method provides a scalable and generalizable approach for solving real-world routing tasks with minimal human intervention, while enabling flexible problem specification through natural language. 
\end{abstract}


\section{INTRODUCTION}
\label{sec:intro}  

Path planning with customized constraints arises in many real-world applications such as logistics, delivery routing, and travel planning. These problems often require finding optimal routes that satisfy both cost objectives and user-specified constraints, such as the number of routes, maximum route length, depot locations, and scheduling requirements. While the Traveling Salesperson Problem (TSP)\cite{christofides2022worst} represents a classical form of route optimization, real-world scenarios frequently involve more complex variations of the vehicle routing problem (VRP)\cite{gutin2006traveling, mor2022vehicle} that significantly increase problem difficulty as new constraints are introduced. Various constrained VRP formulations have been studied, including distance-constrained VRP~\cite{nagarajan2012approximation}, capacitated VRP~\cite{ralphs2003capacitated}, and VRP with time windows~\cite{kallehauge2005vehicle}. While these traditional approaches rely on specialized algorithms and combinatorial optimization techniques, they often require dedicated mathematical formulations, expert knowledge, and extensive tuning for each problem variant, which limits their scalability when new requirements are introduced.

\begin{figure}[htbp]
  \centering
  \includegraphics[width=0.95\linewidth]{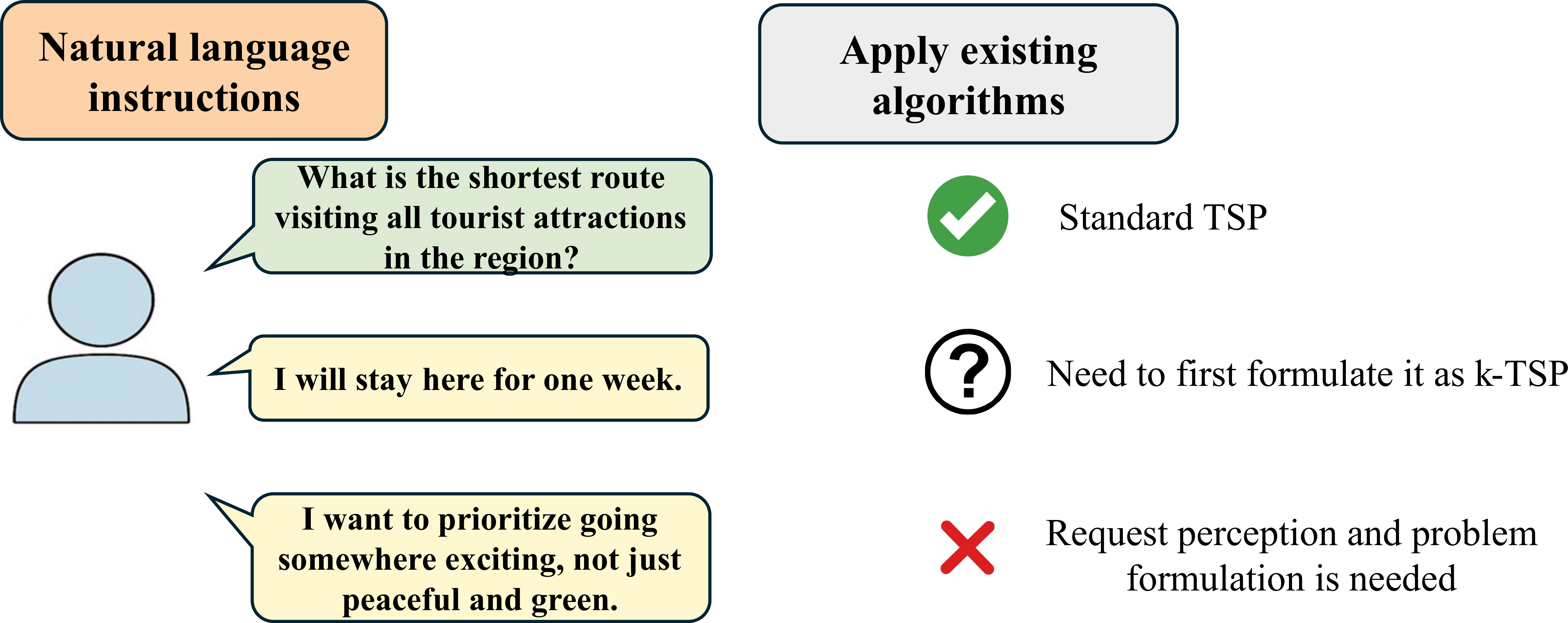}
  \caption{ \label{fig:yourlabel}
  Challenges in applying existing path planning algorithms to practical requests}
\end{figure}

Recent advances in large language models (LLMs) have introduced a promising alternative for solving complex tasks by directly processing natural language input. Trained on massive corpora, LLMs exhibit general-purpose reasoning, planning, and language understanding capabilities. These models have demonstrated strong performance across a wide range of domains, including programming, question answering, data analysis, and mathematical reasoning. Crucially, LLMs can serve as general-purpose problem solvers that allow users to specify tasks in natural language, removing the need for structured problem formulation processes and task-specific optimization algorithms.

In this work, we propose a flexible approach that integrates LLMs with verification and iterative refinement to solve constrained path planning problems directly from natural language input. This approach enables the system to automatically interpret diverse user-defined routing specifications without requiring specialized algorithmic development for each problem variant. Our approach supports both previously studied problem types and novel task specifications by leveraging problem matching and in-context learning~\cite{wies2023learnability}. Through iterative solution generation and verification, the framework ensures that the produced solutions satisfy user-defined constraints while progressively improving solution quality. This combination enables the system to produce feasible and optimized solutions even in cases that were not explicitly formulated during system design.

Our contributions in this research are summarized as follows:
\begin{itemize}
    \item We develop a flexible pipeline that enables large language models to solve constrained path planning problems directly from natural language input, allowing users to specify complex routing tasks conversationally without requiring formal structured problem formulations.
    \item We design a two-pathway system that handles both previously studied problem types via problem matching, and novel task specifications through in-context learning, enabling the system to generalize to new and unseen problem cases.
    \item We integrate an iterative solution generation and verification process, which ensures that generated solutions satisfy user-defined constraints while progressively improving solution quality.
    \item We conduct comprehensive evaluations on a variety of constrained routing scenarios using a state-of-the-art LLM, demonstrating the flexibility, scalability, and practical potential of the proposed approach.
\end{itemize}

\section{Related Work}

There has been extensive prior work on path planning under constraints, including energy, distance, and time, using traditional algorithmic methods~\cite{mor2022vehicle}. For example, an energy-constrained variant requires planned paths to guide robots to visit charging stations before running out of battery power~\cite{wei2018coverage, wei2018log, yu2019coverage}. While approximation algorithms for these problems can offer strong performance guarantee under specific formulations, they often require manual modeling and are less adaptable to new constraint types or combinations. To address the growing need for flexibility, recent research has turned to large language models (LLMs) as a promising alternative. In this section, we review related work that explores how LLMs have been used to perform human-like reasoning and decision-making tasks, followed by studies on guiding or enhancing LLMs for solving various types of optimization problems, including those involving routing and planning.

Previous research involving LLMs has shown how these artificial intelligence systems can be used to perform various tasks. The work in~\cite{tayal2025conversational, neupane2025towards} demonstrated the use of LLMs within the healthcare field to perform various tasks. One way the researchers in~\cite{tayal2025conversational} utilized LLMs was by generating a conversational LLM capable of providing necessary sodium details about various foods to assist heart failure patients in managing their salt consumption. In their work, they utilized OpenAI's text-to-speech and speech-to-text functionalities with a retrieval technique for searching a dataset to find relevant dietary facts about a particular food~\cite{tayal2025conversational}. LLMs have also been shown to automatize various clinical tasks that involve working with private healthcare data while ensuring proper privacy regulations are maintained~\cite{neupane2025towards}. Furthermore, LLMs have also been shown to be capable of assisting and improving education and learning \cite{gupta2025multilingual, thys2025insight}. In~\cite{gupta2025multilingual}, they explored the use of LLMs for performing tasks such as tutoring, grading, and understanding common student mistakes to provide valuable feedback. In addition to assisting students directly, LLMs can be used to provide teachers with feedback on topics students are struggling with to improve a student's education~\cite{thys2025insight}. On a different note, LLMs have also been used to perform other difficult tasks such as creating an ontology as shown by Lippolis \textit{et al.} in~\cite{lippolis2025assessing}. In their work, they introduced a prompt structure with the use of few-shot prompting which was able to help an LLM generate an ontology for various domains. LLMs have been broadly used in a variety of fields to perform various tasks, showing the generalizability of LLMs. 

LLMs have also been utilized to solve optimization problems which can be difficult to solve due to the presence of various constraints and limitations within a problem. One approach implemented by Sartori and Blum~\cite{sartori2025combinatorial} involved using an LLM to enhance the code of a basic optimization algorithm, resulting in a more robust code capable of generating solutions with higher optimality. Their work proved that LLMs are capable at performing challenging tasks, such as improving an optimization algorithm, without the need for human knowledge. Multiple works of research have utilized LLMs to generate heuristics capable of solving various optimization problems \cite{bomer2025leveraging, wang2025planning, zhang2025llm, romera2024mathematical, liu2024evolution, ye2024reevo, liu2024llm4ad, li2025ars}. Wang \textit{et al.} \cite{wang2025planning} proposed a framework capable of improving and generating heuristics without the need for human expertise. In their framework, an LLM is first utilized to analyze an initial heuristic, provide improvement suggestions, then generate a new heuristic containing the suggestions where each heuristic is stored within a search tree. Wang \textit{et al.} also implemented the Monte Carlo Tree Search algorithm iteratively to find the most optimal heuristic generated within the search tree. Furthermore, Li \textit{et al.} \cite{li2025ars} solved various vehicle routing problems (VRP) by utilizing LLMs to generate heuristics capable of adapting to the various constraints present within different VRP variants. Another method used for solving VRPs involved implementing retrieval augmented generation (RAG) to provide the LLM with sample codes and assistive information to improve an LLMs ability to generate code capable of solving a VRP \cite{jiang2025droc}. Zhou \textit{et al.} \cite{zhou2024dynamicroutegpt} proposed a method for solving real-time route selection by utilizing an LLM along with various mathematical models and algorithms including the Markov chain model, Dijkstra algorithm, and Bayesian inference. With this framework, the mathematical models and algorithms provided the LLM with valuable information necessary for the LLM to decide on the next optimal route to take. On another note, the work in \cite{huang2024can} proposed a framework that includes self-debugging and self-verification methods which have been shown to be effective at formulating TSP solutions. Although previous research has implemented the use of external algorithms to enhance an LLMs ability in solving an optimization problem, the researchers in~\cite{huang2024can} solely utilized the capabilities of an LLM for generating the self-debugger and self-verifier utilized in their framework. Alternatively, Elhenawy \textit{et al.} \cite{elhenawy2024visual} tested the use of multimodal large language models (MLLMs) in solving various TSP problems by using visual reasoning only, without the inclusion of any textual information. Their proposed method consists of a multi-agent system containing an initializer agent used to generate an initial route based on visual reasoning, a critic agent which generated multiple improved routes, and finally a scorer agent which assigned a score to each route for an iterative process of returning the highest scored route back to the critic agent for further improvements. While this proposed method worked reliably at refining and evaluating routes, they proposed an alternative method that was more efficient and quicker. Their faster working method consisted of only the initializer and critic agents where they limited the critic agent to only formulate one revised route per iteration as compared to formulating multiple routes for greater efficiency. Many of the methods discussed above included the implementation of an iterative process for refinement or improvement in order to improve an LLM or MLLM's ability to produce optimal solutions~\cite{sartori2025combinatorial, bomer2025leveraging, wang2025planning, huang2024can, elhenawy2024visual}. As this method has shown to be effective in formulating improved optimization solutions, we integrated an iterative process into our proposed system.  

In our work, we presented a system used to improve an LLMs ability to solve various constrained path planning problems described in natural language. In contrast to previously discussed research work, our research works solely on solving naturally written path planning problems as input without the need for initial prompt engineering. Our framework first consists of matching the naturally written routing problem to a corresponding structured path planning prompt obtained from a pre-generated library of various path planning cases, each containing different constraints. On the other hand, if no matching path planning case is found, the LLM self-generates a structured prompt used to assist with the process of solving the inputted problem. Similar to previous research, we incorporate an iterative process used to generate multiple routed solutions to identify the most optimal solution. In contrast to previous work that solved optimization problems by outputting an executable code generated by an LLM~\cite{sartori2025combinatorial, bomer2025leveraging, wang2025planning, huang2024can}, our research aims to utilize an LLM to directly output a routed solution in natural language without the need for code. By implementing LLMs for solving various routing problems in natural language, we prevent the need for a specialized algorithm to solve each variation of a constrained path planning problem. Our proposed framework aims to advance LLMs capabilities for understanding natural language inputs while improving routed solutions through iteration and verification methods.

\begin{figure*}[!t]
\centering
\includegraphics[width=0.95\linewidth]{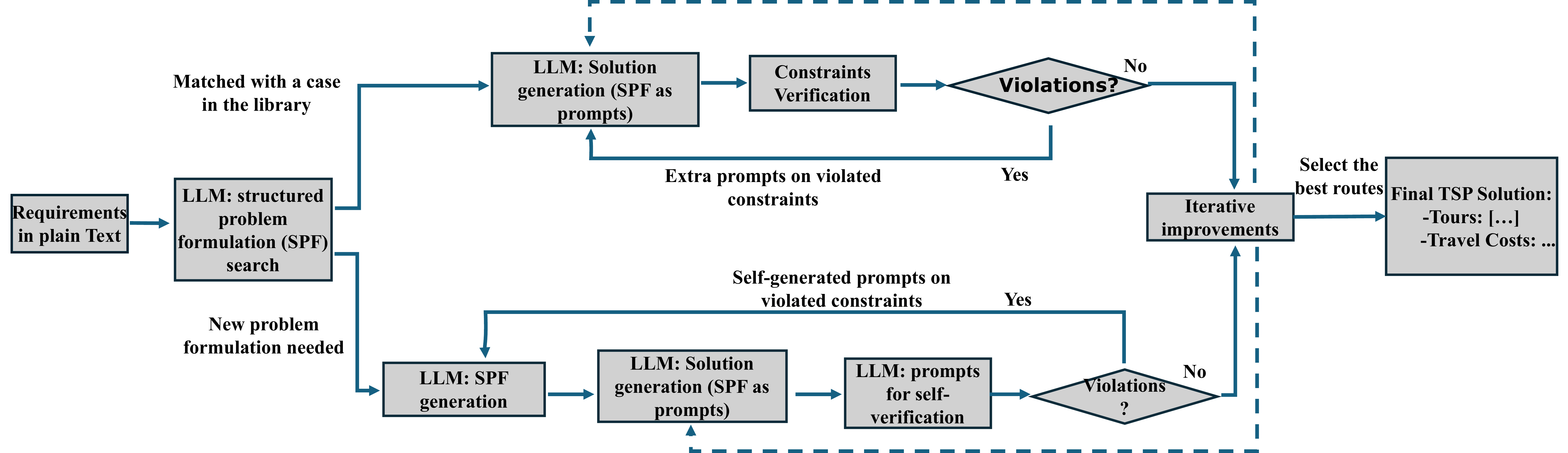}
\caption{ \label{fig:overview}
Overview of a two pathway method capable of solving a natural language path planning problem. An LLM first analyzes the natural language path planning input to find a matching case from a pre-generated path planning case library, then solves the routing problem using a structured prompt from the library. For instances where there is no matching case from the library, the LLM self-generates a structured prompt to solve the problem. Multiple solutions are produced before selecting the most optimal solution as the final output. }
\end{figure*}

\section{Methodology}
In this section, we present a flexible method that enables large language models ({\em LLMs}) to solve constrained path planning problems specified through natural language. The goal of our proposed system is to support flexible, user-friendly problem specifications while ensuring that generated solutions remain feasible and aligned with real-world constraints. By combining natural language understanding with structured solution verification, the system can generalize across problem types and deliver optimized results without requiring manual formulation or domain-specific algorithms.

\subsection{Structured Prompt Formulation for Solving Natural Language Routing Tasks}

Figure~\ref{fig:overview} provides an overview of our system, which includes two pathways for solving constrained path planning problems described in plain text. The upper pathway utilizes an external source to assist the LLM in generating a solution for planning problems that follow known formulations or common constraint patterns. Specifically, we first construct a library that stores multiple path planning problem types, where each entry includes a natural language description along with a corresponding structured problem formulation. Each problem type is parameterized by factors such as the distance matrix between target locations and constraint settings (e.g., number of routes, maximum route length, depot assignments). In addition, each structured formulation includes output format specifications to facilitate consistent evaluation and parsing of planning results.

\begin{figure}[!b]
\centering
\includegraphics[width=0.95\columnwidth]{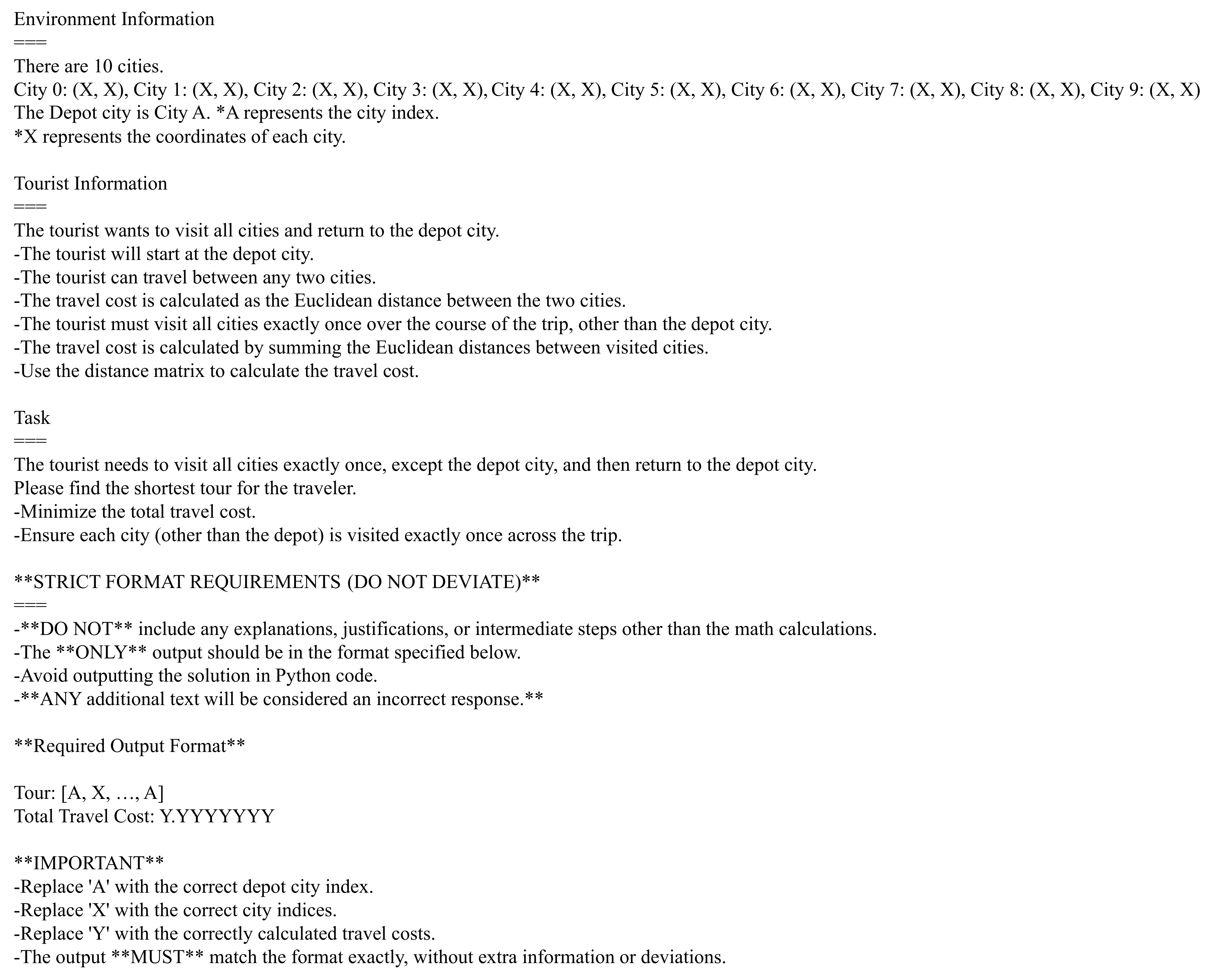}
\caption{ \label{fig:structured_prompt}
A pre-formulated structured prompt example stored within the path planning case library. This prompt is used to assist an LLM in solving the first path planning problem type, the basic TSP problem. }
\end{figure}

Once we receive a plain-text routing request, we pass it to the LLM, which is prompted to interpret the planning constraints and identify the most relevant problem type from our library. Based on the match, the LLM generates a problem case that includes a structured problem formulation (SPF), where each constraint parameter is filled in, along with the optimization objective and required output format. The LLM then uses the SPF to generate a candidate solution. Each constraint in the SPF can be expressed as a mathematical inequality, allowing us to verify the feasibility of the solution by checking whether all constraints are satisfied. Common criteria include ensuring that all required locations are visited, that each non-depot location is visited only once, and that each route starts and ends at a designated depot. If any constraint is violated, we prompt the LLM to revise its solution, explicitly indicating which condition was not met. This generation-verification cycle is repeated until a feasible solution is found or a fixed number of iterations is reached. Once a feasible solution is obtained, we prompt the LLM to optimize it further. In each iteration, the LLM generates a new candidate and is asked to compare it with those from previous rounds. The LLM is encouraged to identify whether the current solution improves upon earlier ones in terms of travel cost. This self-comparison process guides the model toward incremental improvements, resembling the logic of genetic algorithms~\cite{hu2004knowledge}. The best valid solution found across all iterations is returned as the final output.

For new or uncommon planning problems that the LLM cannot match with existing templates, we activate a second pathway in which the LLM generates both the structured problem formulation and the solution in a self-guided manner. To support this, we use single-shot prompting, which involves prompting the LLM with one example to assist the LLM in understanding the given task \cite{zafar2025using}. In our framework, the LLM is prompted with an example from the library consisting of a plain-text routing request and its corresponding structured formulation. Based on this example, the LLM is asked to formulate the problem structure for the new input and then generate an initial solution, following a process similar to the upper pathway. In this case, we cannot apply external mathematical verification since the constraint inequalities are not explicitly known in advance. To improve feasibility, we instead prompt the LLM to perform self-verification. The model is asked to check whether the solution satisfies all constraints described in the generated ({\em SPF}). If a violation is detected, the LLM is prompted to revise the solution by identifying and correcting the specific constraint failure. This self-check-and-correct process continues until a feasible solution is reached or a maximum number of iterations is met. The iterative optimization process that follows mirrors the upper pathway. The LLM is asked to refine its solution by generating new candidates and comparing them to previous ones, selecting the most cost-efficient valid result as the final output.

\subsection{Route Planning Problem Library}
Our framework was tested on four different types of a path planning problem. For each problem type, there are multiple problem cases containing varying city coordinates, depot city indices, and number of travel days. The first problem type consisted of a basic TSP problem. This problem involved generating one route that reduces the total travel cost while visiting each city only once, where the tour starts and ends at a single depot city. Figure~\ref{fig:structured_prompt} shows a pre-formulated structured prompt of our first path planning problem type stored within our case library. The second problem type involved generating multiple routes for a multi-day trip. In this case, there is a single depot city that each tour must start and end at. Additionally, the non-depot cities must only be visited once throughout the entire trip where the total travel cost is minimized. Our third problem type similarly consisted of generating multiple routes for a multi-day trip, however, each new travel day has a new depot city, where the tours for each day must start and end at its assigned depot city. The constraints for this case also involved minimizing the total travel cost while also visiting the non-depot cities only once throughout the entire trip. For each problem type described above, a corresponding structured prompt was formulated and saved to the case library. To assist an LLM in solving a natural language path-planning problem, the structured prompts contain detailed information about the problem constraints, objective, and specific format requirements for the output. Additionally, a distance matrix containing the euclidean distances between each city, is provided to the LLM. A fourth problem type was created which consisted of various path planning problems, each with different constraints. These routing problems are tested along the self-solving pathway where a structured prompt is generated by the LLM rather than generating a prompt before hand. 

\begin{table}[!b]
\centering
\caption{Average cost with and without iteration, average reduction in cost, and refinement success rate for each problem type}
\label{tab:metrics}
\begin{tabular}{|c|c|c|c|c|}
\hline
\makecell{\textbf{Problem Type}} &
\makecell{\textbf{Without Iteration}} &
\makecell{\textbf{With Iteration}} &
\makecell{\textbf{Average Cost}\\\textbf{Reduction}} &
\makecell{\textbf{Refinement Success}\\\textbf{Rate}} \\
\hline
1 & 287.87 & 272.66 & 5.28\% & 62\% \\
\hline 
2 & 589.45 & 537.52 & 8.81\% & 78\% \\
\hline
3 & 328.13 & 299.31 & 8.78\% & 50\% \\
\hline 
4 & 460.21 & 428.23 & 6.95\% & 66\% \\
\hline
\end{tabular}
\end{table}

\subsection{Large Language Models}
Our proposed method was tested on various LLMs, including Llama3.3 70b, GPT-4, GPT-4-Turbo, and GPT-4o. For our open-source model, we chose the Llama3.3 70b model due to its high performance and various capabilities, including its ability to solve mathematical problems using heuristics, perform high-level question and answering, and understand multiple languages \cite{zafar2025using}. To evaluate our proposed system and obtain our numerical results, we utilized the GPT-4-Turbo model since this LLM has proved to exceed the performance and capabilities of other LLMs when comparing their performance on various types of tasks \cite{chacon2024large, zhang2023using, lakhlani2025evaluating}. 

\begin{figure}[!t]
  \centering
  \begin{tabular}{@{}c@{\hspace{0.04\linewidth}}c@{}}
    \includegraphics[width=0.45\linewidth]{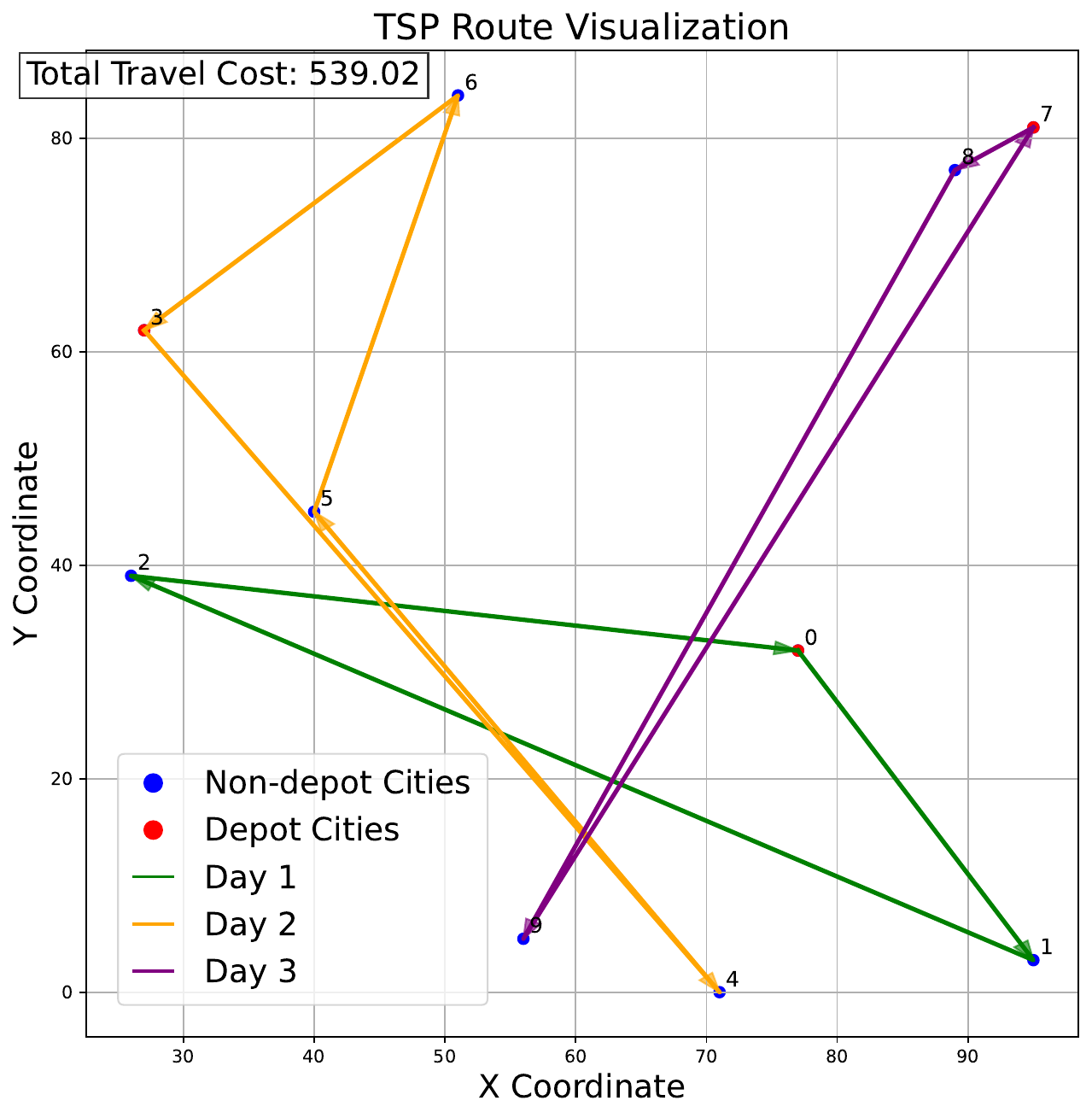} &
    \includegraphics[width=0.45\linewidth]{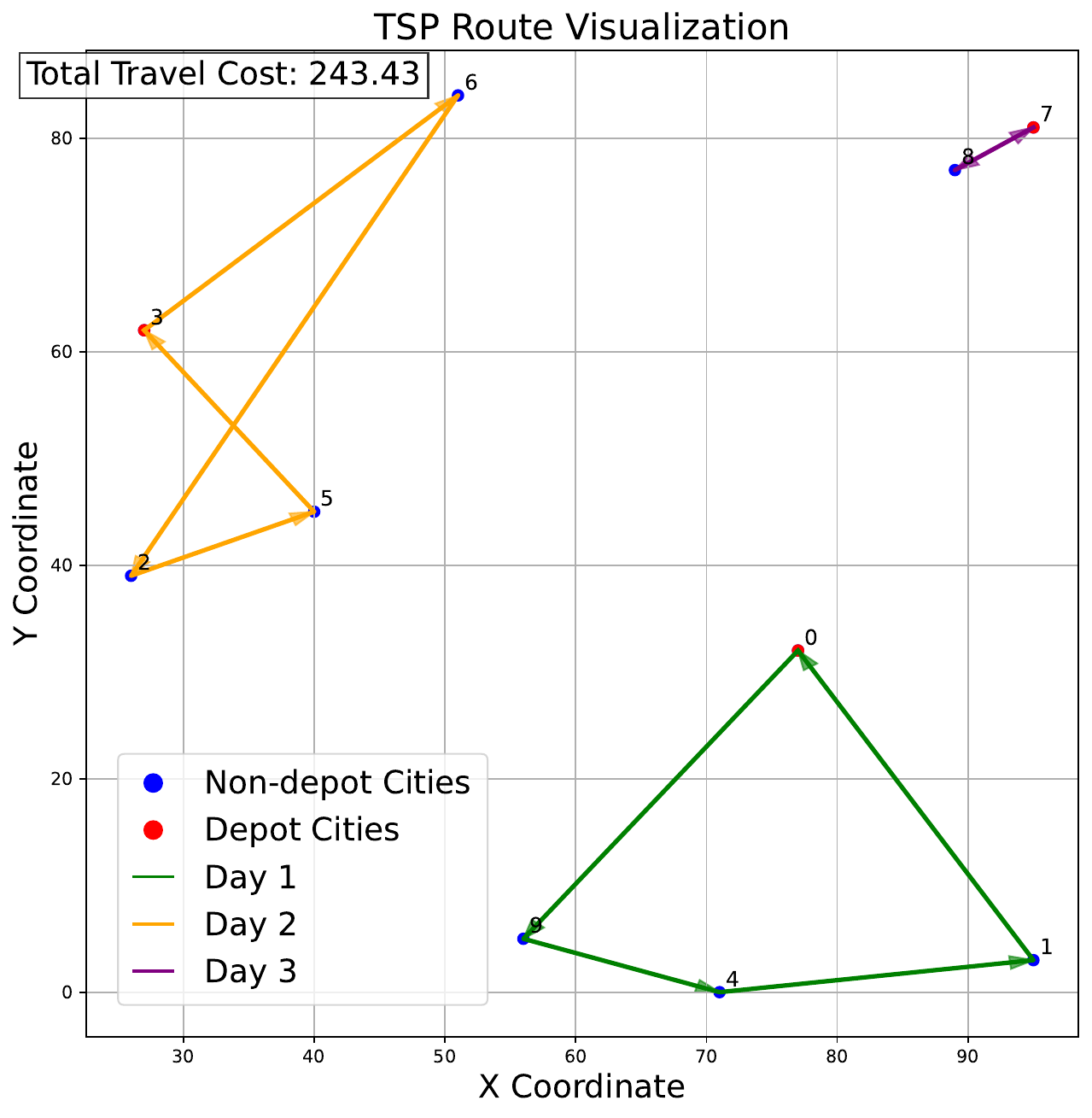} \\
    (a) Without iterative refinement & (b) With iterative refinement \\
  \end{tabular}
  \caption{ \label{fig:side-by-side}
  Solutions for a three-day trip to 10 cities with distinct depot cities per day (problem type 3). (a) shows a suboptimal route generated without iteration. (b) shows a better route obtained using iterative refinement.}
\end{figure}

\section{Experiment Results}

In this section, we present experimental results to evaluate the effectiveness of our proposed system for solving constrained path planning problems using large language models. Our evaluation focuses on two core components: the ability of the LLM to perform reliable self-verification based on the structured problem formulation (SPF), and the impact of the self-iterative refinement module on solution quality. We conduct experiments across all defined planning problem types, demonstrating how the framework handles varied constraints and objectives. In addition to quantitative evaluation of feasibility and cost optimality, we present representative planning results to illustrate the system's behavior in practice.

We first compare the performance of the LLM with and without self-verification in terms of solution feasibility. When running an LLM 800 times to formulate planned routes to the constrained path planning problems with the use of self-verification, 95.38\% of the solutions produced were valid solutions which accurately followed the problem constraints. In addition, we omitted the verification module and tested the same constrained path planning problems for consistency, which resulted in 85.63\% of valid solutions. These results validate that our verification module was beneficial for improving an LLMs ability to understand and follow the requirements of path planning problems provided in natural language.

We then evaluate how iterative refinement influences solution quality by measuring total travel cost. Travel cost was calculated as the sum of the euclidean distances between each visited city along the planned routes. Within our proposed system, the iterative process, along with the solution improvement prompt, aimed to reduce the path cost of the output solution. When testing our system with and without this iterative portion on each constrained routing problem, we found that 64\% of the solutions had reduction in path costs. This signifies that the iterative portion of our pipeline is effective in optimizing the solutions. Furthermore, we obtained the average total travel costs for each problem type with and without the use of the iterative aspect, which is shown in Table~\ref{tab:metrics}. When iteration is implemented within our system, the average total travel costs are lower for each case as compared to the average costs when iteration is omitted. This proves that the iterative section of our system helps with increasing the optimality of the routed solutions produced by an LLM. From the illustration shown in Figure~\ref{fig:side-by-side}, we can see how our proposed method is capable of improving an LLMs ability to produce better solutions to a constrained path planning problem.

Additionally, we assessed an LLMs response with and without the use of formulated structured prompts (i.e. structured problem formulation {\em SPF}) to demonstrate its importance for enabling self verification. Without {\em SPF}, we can only rely on general prompts like 'check the solution against the constraints' for self-verification. As already been mentioned in this section, when implementing either pre-formulated structured prompts or LLM-generated structured prompts, we noticed that LLMs were able to successfully produce routes while accurately following the constraints of each natural language routing problem. Another benefit of the structured prompts from the case library is the consistent output format for the solutions, favoring further processing. In contrast, without {\em SPF}, LLMs struggled to provide valid and easy-to-interpret solutions. Since LLMs are trained on mainly text data, LLMs naturally produce outputs in a descriptive human language text, making it hard to evaluate the success rate at a large scale. We manually examined 5 cases and all the outputs violate given constraints. 

\begin{figure}[!b]
  \centering
  \begin{tabular}{@{}c@{\hspace{0.04\linewidth}}c@{}}
    \includegraphics[width=0.45\linewidth]{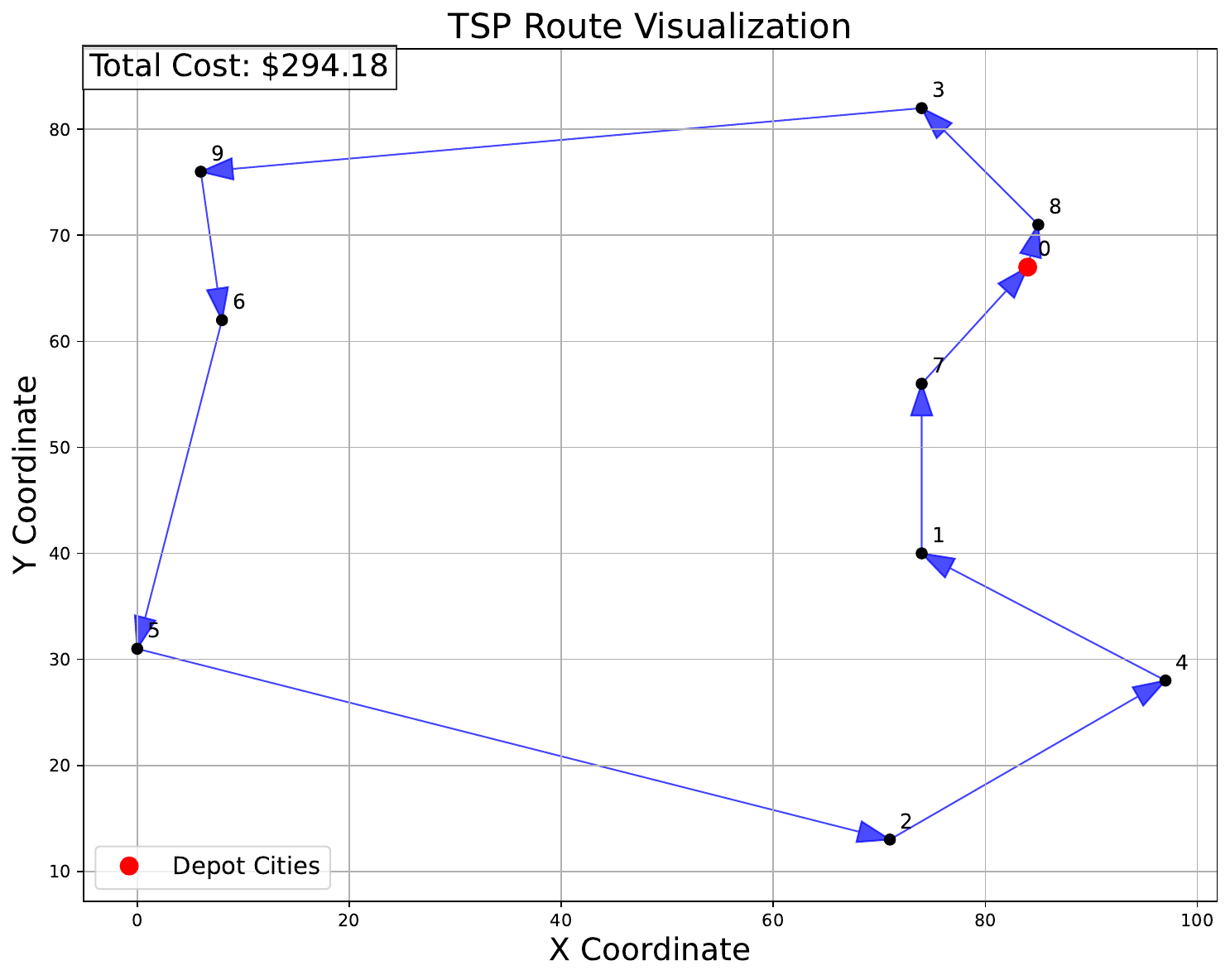} &
    \includegraphics[width=0.45\linewidth]{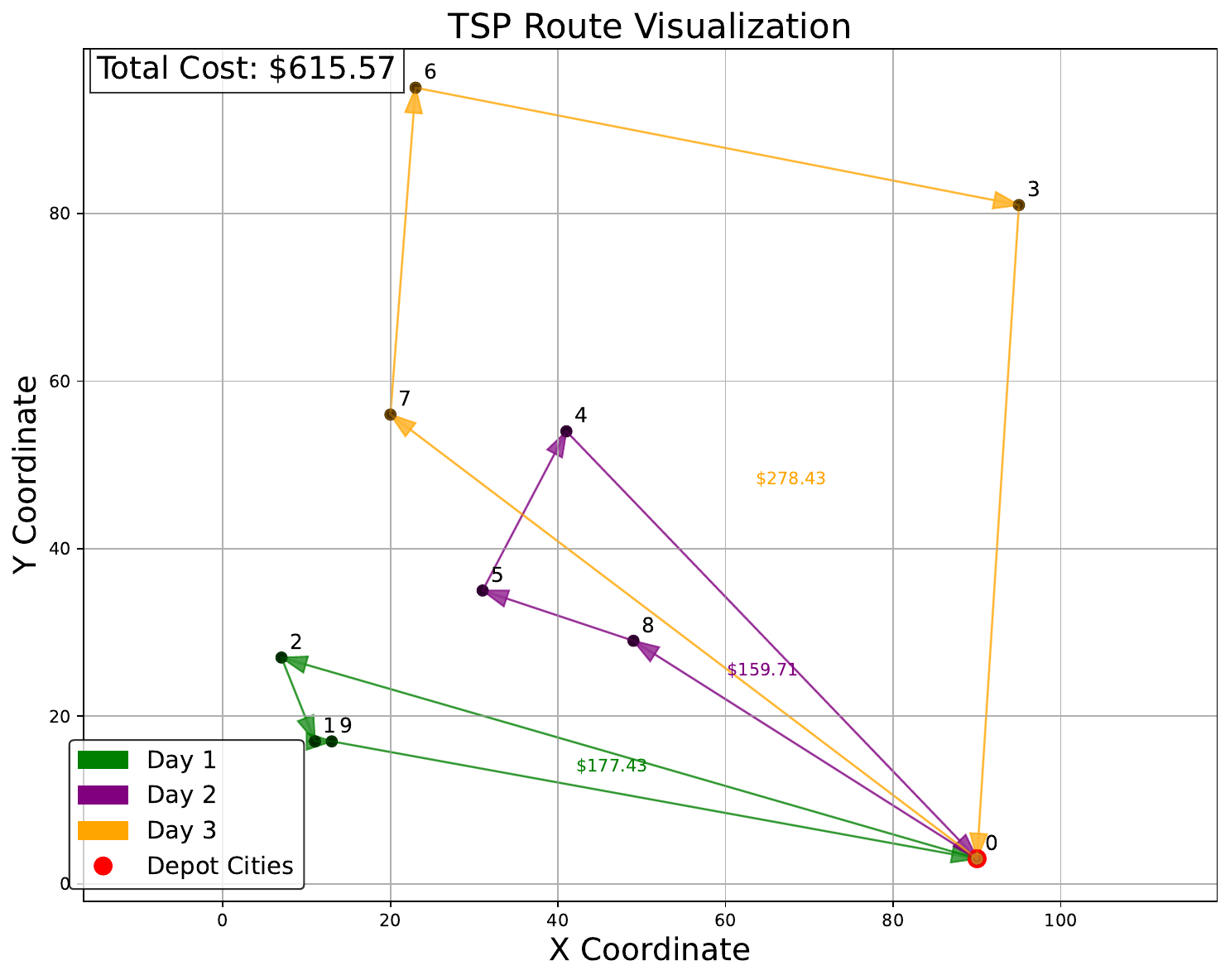} \\
    \includegraphics[width=0.45\linewidth]{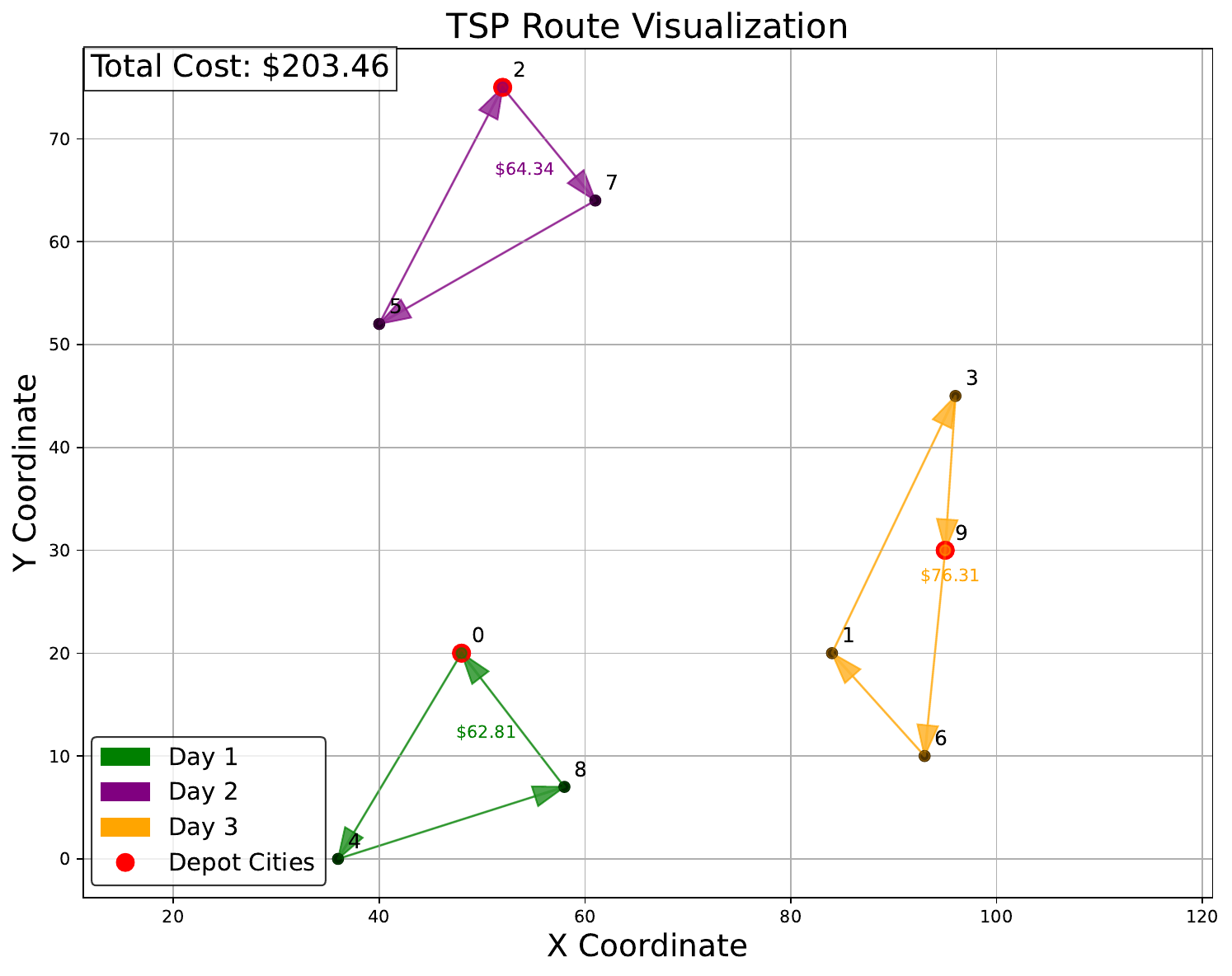} &
    \includegraphics[width=0.45\linewidth]{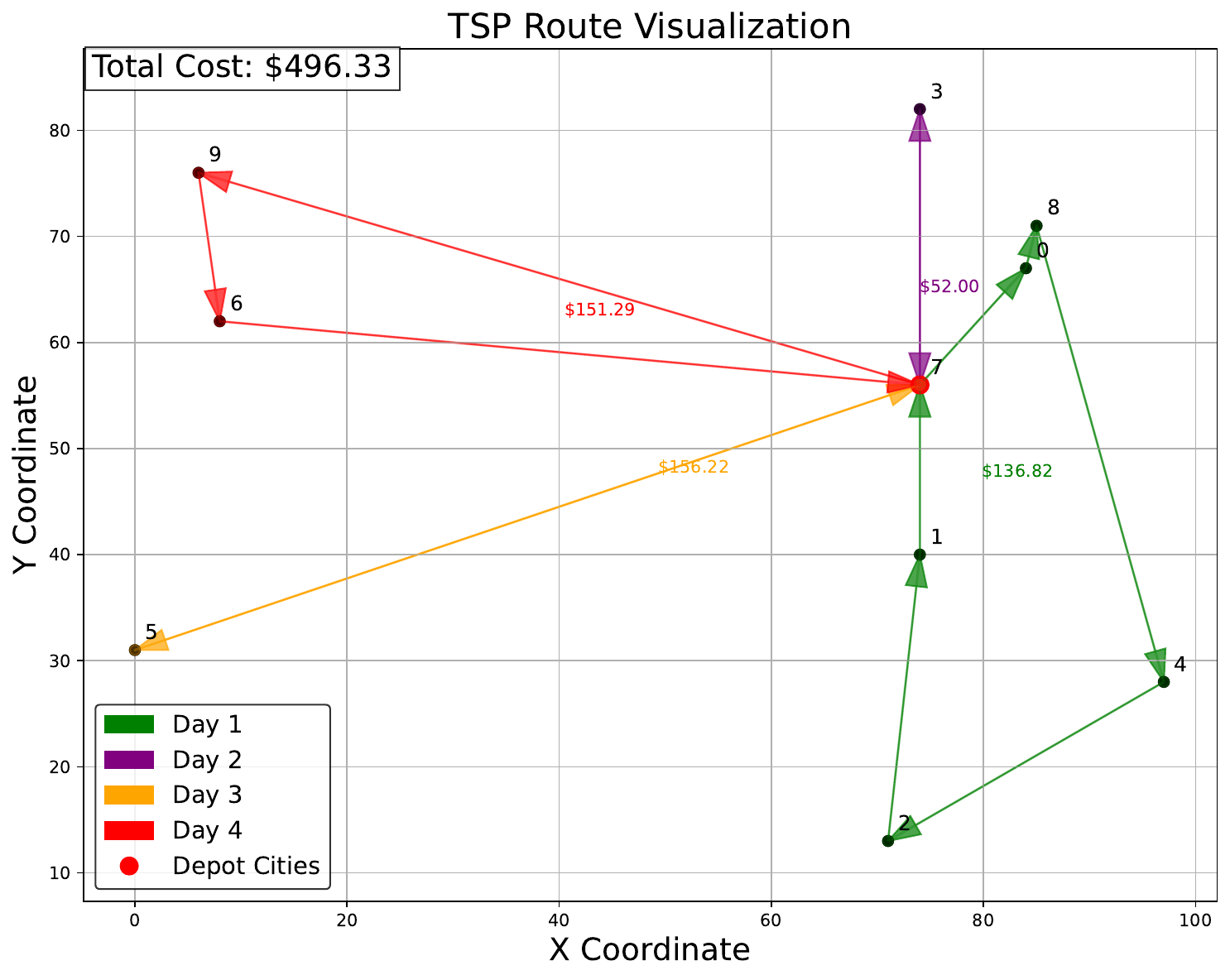} \\
  \end{tabular}
  \caption{ \label{fig:grid_images}
  Example routed solutions for four constrained path-planning problem types generated by an LLM: problem type 1 (top left), problem type 2 (top right), problem type 3 (bottom left), and problem type 4 (bottom right) are highly optimal routes with corresponding travel costs for a problem case of each problem type.}
\end{figure}

Finally, we present representative planning results across all problem types to illustrate how the system performs under different objectives and constraints. Figure~\ref{fig:grid_images} shows the routed solutions of one problem case for the four problem types we tested. As we can see, our system was able to accurately ensure an LLM provided valid routes that properly complied to the constraints of each problem type. Additionally, each LLM was guided to output an optimal yet efficient and cost effective travel route for each routing problem containing its own set of constraints. Overall, this shows the effectiveness of our proposed system in assisting LLMs to correctly solve and find optimal solutions to various path planning problems. 

\subsection{Limitations of Current LLM-based Approach}
In this section, we report two limitations of current approach: theoretical performance guarantee, and scalability.

Despite the success in improving 
{\em LLMs} capability to plan feasible and optimized paths under constraints described by plain text, there is no performance guarantee on the solutions generated them. All the four problem types in this paper are NP-hard, which means no known polynomial-time algorithms can solve the problem optimally unless P = NP. {\em LLMs} can neither prove that the solutions are generated by approximation algorithms to guarantee that $Cost_{sol} \leq \alpha Cost_{opt}$, where $\alpha$ is the approximation rate.

A second limitation we observed with current {\em LLM-based} approaches is scalability: performance tends to degrade as the number of target locations in the input increases. In our experiments, we found that most models struggled to generate valid solutions when the problem involved more than 15–20 cities. When this limit was reached or exceeded, the LLMs often produced invalid outputs, including missing cities, repeated visits to the same location, incorrect depot assignments, or hallucinated responses that did not follow the required output format. Notably, when such errors occurred, our self-verification module was still able to detect constraint violations and notify the user, even though the {\em LLM} itself was unable to reliably correct the issues. This highlights the importance of incorporating {\em SPF} and self-verification mechanisms to ensure solution validity.

\section {Conclusion and Future Work}

In conclusion, we presented an {\em LLM-based} approach for solving a variety of path planning problems where constraints are specified in natural language. Our system leverages a case library containing structured problem formulations ({\em SPF}) to guide the LLM in interpreting and solving planning problems under different constraint types. These formulations are integrated with a self-verification module that checks the feasibility of candidate solutions against the specified constraints. To further improve solution quality, we introduce an iterative refinement process in which the LLM generates multiple solutions and selects the one with the lowest total cost. While our approach demonstrates strong performance on problems with up to 20 locations, current LLMs exhibit limitations in scalability. As future work, we plan to enhance the framework’s ability to handle larger and more complex instances. One promising direction is a hybrid approach, where we utilize {\em LLMs} to further convert the {\em SPF} into an integer linear programming ({\em ILP})~\cite{kulkarni1985integer} case and delegate the final optimization step to a conventional {\em ILP} solver.

\section{Acknowledgment}
This work was supported by NSF~2452203, and the Special License Plate funds granted through the Harbor Branch Oceanographic Institute Foundation, FAU. 

\bibliography{ref} 

@article{tayal2025conversational,
  title={Conversational Assistants to support Heart Failure Patients: comparing a Neurosymbolic Architecture with ChatGPT},
  author={Tayal, Anuja and Salunke, Devika and Di Eugenio, Barbara and Allen-Meares, Paula and Abril, Eulalia Puig and Garcia, Olga and Dickens, Carolyn and Boyd, Andrew},
  journal={arXiv preprint arXiv:2504.17753},
  year={2025}
}

@article{gupta2025multilingual,
  title={Multilingual performance biases of large language models in education},
  author={Gupta, Vansh and Chowdhury, Sankalan Pal and Zouhar, Vil{\'e}m and Rooein, Donya and Sachan, Mrinmaya},
  journal={arXiv preprint arXiv:2504.17720},
  year={2025}
}

@article{thys2025insight,
  title={INSIGHT: Bridging the Student-Teacher Gap in Times of Large Language Models},
  author={Thys, Jarne and Vanbrabant, Sebe and Vanacken, Davy and Ruiz, Gustavo Rovelo},
  journal={arXiv preprint arXiv:2504.17677},
  year={2025}
}

@article{neupane2025towards,
  title={Towards a hipaa compliant agentic ai system in healthcare},
  author={Neupane, Subash and Mittal, Sudip and Rahimi, Shahram},
  journal={arXiv preprint arXiv:2504.17669},
  year={2025}
}

@article{lippolis2025assessing,
  title={Assessing the Capability of Large Language Models for Domain-Specific Ontology Generation},
  author={Lippolis, Anna Sofia and Saeedizade, Mohammad Javad and Keskisarkka, Robin and Gangemi, Aldo and Blomqvist, Eva and Nuzzolese, Andrea Giovanni},
  journal={arXiv preprint arXiv:2504.17402},
  year={2025}
}

@article{sartori2025combinatorial,
  title={Combinatorial Optimization for All: Using LLMs to Aid Non-Experts in Improving Optimization Algorithms},
  author={Sartori, Camilo Chac{\'o}n and Blum, Christian},
  journal={arXiv preprint arXiv:2503.10968},
  year={2025}
}

@article{bomer2025leveraging,
  title={Leveraging large language models to develop heuristics for emerging optimization problems},
  author={B{\"o}mer, Thomas and Koltermann, Nico and Disselnmeyer, Max and D{\"o}rr, Laura and Meyer, Anne},
  journal={arXiv preprint arXiv:2503.03350},
  year={2025}
}

@article{wang2025planning,
  title={Planning of Heuristics: Strategic Planning on Large Language Models with Monte Carlo Tree Search for Automating Heuristic Optimization},
  author={Wang, Hui and Zhang, Xufeng and Mu, Chaoxu},
  journal={arXiv preprint arXiv:2502.11422},
  year={2025}
}

@article{huang2024can,
  title={Can Large Language Models Solve Robot Routing?},
  author={Huang, Zhehui and Shi, Guangyao and Sukhatme, Gaurav S},
  journal={arXiv preprint arXiv:2403.10795},
  year={2024}
}

@article{elhenawy2024visual,
  title={Visual reasoning and multi-agent approach in multimodal large language models (mllms): Solving tsp and mtsp combinatorial challenges},
  author={Elhenawy, Mohammed and Abutahoun, Ahmad and Alhadidi, Taqwa I and Jaber, Ahmed and Ashqar, Huthaifa I and Jaradat, Shadi and Abdelhay, Ahmed and Glaser, Sebastien and Rakotonirainy, Andry},
  journal={arXiv preprint arXiv:2407.00092},
  year={2024}
}

@article{mor2022vehicle,
  title={Vehicle routing problems over time: a survey},
  author={Mor, Andrea and Speranza, Maria Grazia},
  journal={Annals of Operations Research},
  volume={314},
  number={1},
  pages={255--275},
  year={2022},
  publisher={Springer}
}

@inproceedings{wei2018coverage,
  title={Coverage path planning under the energy constraint},
  author={Wei, Minghan and Isler, Volkan},
  booktitle={2018 IEEE International Conference on Robotics and Automation (ICRA)},
  pages={368--373},
  year={2018},
  organization={IEEE}
}

@inproceedings{wei2018log,
  title={A log-approximation for coverage path planning with the energy constraint},
  author={Wei, Minghan and Isler, Volkan},
  booktitle={Proceedings of the International Conference on Automated Planning and Scheduling},
  volume={28},
  pages={532--539},
  year={2018}
}

@book{kallehauge2005vehicle,
  title={Vehicle routing problem with time windows},
  author={Kallehauge, Brian and Larsen, Jesper and Madsen, Oli BG and Solomon, Marius M},
  year={2005},
  publisher={Springer}
}

@article{nagarajan2012approximation,
  title={Approximation algorithms for distance constrained vehicle routing problems},
  author={Nagarajan, Viswanath and Ravi, R},
  journal={Networks},
  volume={59},
  number={2},
  pages={209--214},
  year={2012},
  publisher={Wiley Online Library}
}

@article{ralphs2003capacitated,
  title={On the capacitated vehicle routing problem},
  author={Ralphs, Ted K and Kopman, Leonid and Pulleyblank, William R and Trotter, Leslie E},
  journal={Mathematical programming},
  volume={94},
  pages={343--359},
  year={2003},
  publisher={Springer}
}

@book{gutin2006traveling,
  title={The traveling salesman problem and its variations},
  author={Gutin, Gregory and Punnen, Abraham P},
  volume={12},
  year={2006},
  publisher={Springer Science \& Business Media}
}

@article{christofides2022worst,
  title={Worst-case analysis of a new heuristic for the travelling salesman problem},
  author={Christofides, Nicos},
  journal={Operations Research Forum},
  volume={3},
  number={1},
  pages={20},
  year={2022},
  publisher={Springer}
}

@inproceedings{hu2004knowledge,
  title={A knowledge based genetic algorithm for path planning of a mobile robot},
  author={Hu, Yanrong and Yang, Simon X},
  booktitle={IEEE International Conference on Robotics and Automation, 2004. Proceedings. ICRA'04. 2004},
  volume={5},
  pages={4350--4355},
  year={2004},
  organization={IEEE}
}

@inproceedings{zafar2025using,
  title={Using Large Language Models to Solve the Electric Vehicle Routing Problem with Advanced Prompting Techniques},
  author={Zafar, Usman and Bayhan, Sertac},
  booktitle={2025 IEEE 19th International Conference on Compatibility, Power Electronics and Power Engineering (CPE-POWERENG)},
  pages={1--6},
  year={2025},
  organization={IEEE}
}

@article{kulkarni1985integer,
  title={Integer programming formulations of vehicle routing problems},
  author={Kulkarni, RV and Bhave, Pramod R},
  journal={European journal of operational research},
  volume={20},
  number={1},
  pages={58--67},
  year={1985},
  publisher={Elsevier}
}

@article{zhang2025llm,
  title={LLM-Driven Instance-Specific Heuristic Generation and Selection},
  author={Zhang, Shaofeng and Liu, Shengcai and Lu, Ning and Wu, Jiahao and Liu, Ji and Ong, Yew-Soon and Tang, Ke},
  journal={arXiv preprint arXiv:2506.00490},
  year={2025}
}

@article{romera2024mathematical,
  title={Mathematical discoveries from program search with large language models},
  author={Romera-Paredes, Bernardino and Barekatain, Mohammadamin and Novikov, Alexander and Balog, Matej and Kumar, M Pawan and Dupont, Emilien and Ruiz, Francisco JR and Ellenberg, Jordan S and Wang, Pengming and Fawzi, Omar and others},
  journal={Nature},
  volume={625},
  number={7995},
  pages={468--475},
  year={2024},
  publisher={Nature Publishing Group UK London}
}

@article{liu2024evolution,
  title={Evolution of heuristics: Towards efficient automatic algorithm design using large language model},
  author={Liu, Fei and Tong, Xialiang and Yuan, Mingxuan and Lin, Xi and Luo, Fu and Wang, Zhenkun and Lu, Zhichao and Zhang, Qingfu},
  journal={arXiv preprint arXiv:2401.02051},
  year={2024}
}

@article{ye2024reevo,
  title={Reevo: Large language models as hyper-heuristics with reflective evolution},
  author={Ye, Haoran and Wang, Jiarui and Cao, Zhiguang and Berto, Federico and Hua, Chuanbo and Kim, Haeyeon and Park, Jinkyoo and Song, Guojie},
  journal={Advances in neural information processing systems},
  volume={37},
  pages={43571--43608},
  year={2024}
}

@article{liu2024llm4ad,
  title={Llm4ad: A platform for algorithm design with large language model},
  author={Liu, Fei and Zhang, Rui and Xie, Zhuoliang and Sun, Rui and Li, Kai and Lin, Xi and Wang, Zhenkun and Lu, Zhichao and Zhang, Qingfu},
  journal={arXiv preprint arXiv:2412.17287},
  year={2024}
}

@inproceedings{jiang2025droc,
  title={DRoC: Elevating large language models for complex vehicle routing via decomposed retrieval of constraints},
  author={Jiang, Xia and Wu, Yaoxin and Zhang, Chenhao and Zhang, Yingqian},
  booktitle={13th international Conference on Learning Representations, ICLR 2025},
  year={2025},
  organization={OpenReview. net}
}

@article{li2025ars,
  title={ARS: Automatic Routing Solver with Large Language Models},
  author={Li, Kai and Liu, Fei and Wang, Zhenkun and Tong, Xialiang and Han, Xiongwei and Yuan, Mingxuan and Zhang, Qingfu},
  journal={arXiv preprint arXiv:2502.15359},
  year={2025}
}

@article{zhou2024dynamicroutegpt,
  title={Dynamicroutegpt: A real-time multi-vehicle dynamic navigation framework based on large language models},
  author={Zhou, Ziai and Zhou, Bin and Liu, Hao},
  journal={arXiv preprint arXiv:2408.14185},
  year={2024}
}

@inproceedings{chacon2024large,
  title={Large language models for the automated analysis of optimization algorithms},
  author={Chac{\'o}n Sartori, Camilo and Blum, Christian and Ochoa, Gabriela},
  booktitle={Proceedings of the Genetic and Evolutionary Computation Conference},
  pages={160--168},
  year={2024}
}

@article{zhang2023using,
  title={Using large language models for hyperparameter optimization},
  author={Zhang, Michael R and Desai, Nishkrit and Bae, Juhan and Lorraine, Jonathan and Ba, Jimmy},
  journal={arXiv preprint arXiv:2312.04528},
  year={2023}
}

@article{lakhlani2025evaluating,
  title={Evaluating large language models for surgical chart review of second stage implant-based breast reconstruction: a comparative analysis of manual review, GPT-3.5 Turbo, and GPT-4 Turbo},
  author={Lakhlani, Devi and Dadhania, Dhruv and Nazerali, Rahim},
  journal={European Journal of Plastic Surgery},
  volume={48},
  number={1},
  pages={23},
  year={2025},
  publisher={Springer}
}

@article{wies2023learnability,
  title={The learnability of in-context learning},
  author={Wies, Noam and Levine, Yoav and Shashua, Amnon},
  journal={Advances in Neural Information Processing Systems},
  volume={36},
  pages={36637--36651},
  year={2023}
}

@inproceedings{yu2019coverage,
  title={Coverage of an environment using energy-constrained unmanned aerial vehicles},
  author={Yu, Kevin and O’Kane, Jason M and Tokekar, Pratap},
  booktitle={2019 international conference on robotics and automation (ICRA)},
  pages={3259--3265},
  year={2019},
  organization={IEEE}
}
\bibliographystyle{spiebib} 

\end{document}